\documentclass[10pt,twocolumn,letterpaper]{article}

\usepackage{wacv}
\usepackage{times}
\usepackage{epsfig}
\usepackage{graphicx}
\usepackage{amsmath}
\usepackage{amssymb}
\usepackage{booktabs}
\usepackage{subfigure}
\usepackage{algorithm}
\usepackage{algorithmic}
\usepackage{wrapfig}
\usepackage{multirow}
\usepackage{xcolor}
\usepackage{stackengine}

\newcommand{\myparagraph}[1]{\noindent\textbf{#1}}
\newcommand*{\Scale}[2][4]{\scalebox{#1}{$#2$}}%

\renewcommand{\algorithmiccomment}[1]{\bgroup\hfill//~#1\egroup}

\def\wacvPaperID{699} 

\wacvapplicationstrack 

\wacvfinalcopy 

\ifwacvfinal
\usepackage[breaklinks=true,bookmarks=false]{hyperref}
\else
\usepackage[pagebackref=true,breaklinks=true,colorlinks,bookmarks=false]{hyperref}
\fi

\begin{document}

\title{Deep Statistic Shape Model for Myocardium Segmentation}

\author{Xiaoling Hu\thanks{Work done during internship at United Imaging Intelligence America.}\\
Stony Brook University\\
{\tt\small xiaolhu@cs.stonybrook.edu}
\and
Xiao Chen, Yikang Liu, Eric Z. Chen, Terrence Chen, Shanhui Sun\thanks{Shanhui Sun is the corresponding author.}\\
United Imaging Intelligence America\\
{\tt\small \{xiao.chen01, yikang.liu, zhang.chen, terrence.chen, shanhui.sun\}@uii-ai.com}
}

\maketitle
\thispagestyle{empty}

\begin{abstract}
Accurate segmentation and motion estimation of myocardium have always been important in clinic field, which essentially contribute to the downstream diagnosis. However, existing methods cannot always guarantee the shape integrity for myocardium segmentation. In addition, motion estimation requires point correspondence on the myocardium region across different frames. In this paper, we propose a novel end-to-end deep statistic shape model to focus on myocardium segmentation with both shape integrity and boundary correspondence preserving. Specifically, myocardium shapes are represented by a fixed number of points, whose variations are extracted by Principal Component Analysis (PCA). Deep neural network is used to predict the transformation parameters (both affine and deformation), which are then used to warp the mean point cloud to the image domain. Furthermore, a differentiable rendering layer is introduced to incorporate mask supervision into the framework to learn more accurate point clouds. In this way, the proposed method is able to consistently produce anatomically reasonable segmentation mask without post processing. Additionally, the predicted point cloud guarantees boundary correspondence for sequential images, which contributes to the downstream tasks, such as the motion estimation of myocardium. We conduct several experiments to demonstrate the effectiveness of the proposed method on several benchmark datasets.
\end{abstract}

\section{Introduction}

\begin{figure*}[t]
\centering
\subfigure[Ori]{
\includegraphics[width=0.14\textwidth, angle =180]{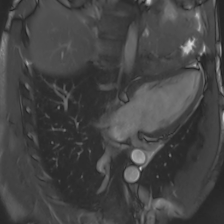}}
\hspace{-.08in}
\subfigure[GT]{
\includegraphics[width=0.14\textwidth, angle =180]{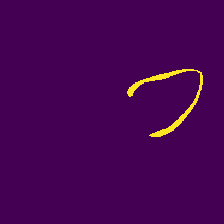}}
\hspace{-.08in}
\subfigure[Point]{
\includegraphics[width=0.14\textwidth, angle =180]{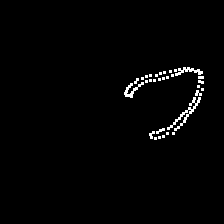}}
\hspace{-.08in}
\subfigure[U-Net]{
\includegraphics[width=0.14\textwidth, angle =180]{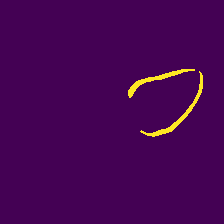}}
\hspace{-.08in}
\subfigure[Our Msk]{
\includegraphics[width=0.14\textwidth, angle =180]{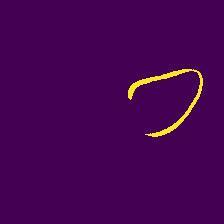}}
\hspace{-.08in}
\subfigure[Our Point]{
\includegraphics[width=0.14\textwidth, angle =180]{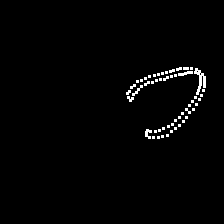}}

\caption{An illustration of the proposed method. The task here is to segment the left ventricle myocardium on the long-axis view. From left to right: (\textbf{a}) original image, (\textbf{b}) gt mask, (\textbf{c}) gt point clouds, (\textbf{d}) mask predicted by U-Net, (\textbf{e}) mask and (\textbf{f}) point clouds predicted by the proposed method. Though U-Net achieves high pixel accuracy (Dice: 0.90), it is prone to shape errors.}
\label{teaser}
\end{figure*}

Myocardium (especially left ventricle, LV) segmentation and motion estimation in medical images such as cardiac Magnetic Resonance Imaging (MRI) are the cornerstones of many computer-aided diagnosis. For example, myocardium wall motion, an important heart function measurement, requires the segmentation of the heart muscle and tracking of its motion through time. Fig.~\ref{teaser}(b) shows LV myocardium mask and Fig.~\ref{teaser}(c) shows myocardium wall (point cloud) delineated on the original cardiac MRI image Fig.~\ref{teaser}(a). The segmentation task can be performed using convolutional neural networks (CNNs). In the field of computer vision, CNNs have demonstrated its power in the image segmentation~\cite{long2015fully,he2017mask,chen2017deeplab,chen2018deeplab,deeplabv32018}. However, since deep learning models are optimized via pixel-wise loss functions, such as mean square error loss (MSE) and cross entropy loss, they usually miss explicit shape constraints. See Fig.~\ref{teaser} as an illustration. U-Net~\cite{ronneberger2015u} is one of the most powerful segmentation frameworks, especially for biomedical images with fine-scale structures. As Fig.~\ref{teaser}(d) shows, though U-Net achieves quite satisfactory pixel accuracy (Dice: 0.90), \textit{it is prone to shape errors such as broken connections}. In addition, as these methods optimize the pixel-wise loss for each pixel independently, the produced masks are not able to preserve boundary correspondence that is required in motion estimation.

Active Shape Models (ASM)~\cite{cootes1995active,yan2010discrete} have been proposed to deal with the shape integrity issue. Specifically, a fixed number of points are used to represent shapes, and then PCA is adopted to establish point distribution models (PDM)~\cite{cootes2004statistical} by finding the variations of the shapes across the training dataset. In the inference stage, a shape in agreement with the shape model will be generated by feeding both the test image and the shape prior into the established shape model. The generated shape is guaranteed with shape integrity, regardless of training data size/image quality. 

Researchers have also incorporated shape priors into deep learning frameworks~\cite{duan2019automatic,zotti2018convolutional,oktay2017anatomically,painchaud2019cardiac,yue2019cardiac}. Instead of using handcrafted features, CNNs are used to extract complex appearance features from images to learn the shape parameters of the statistical shape  models~\cite{milletari2017integrating,attar20193d,tilborghs2020shape}. These methods either learn the shape parameters and then produce the final segmentation as post-processing, or directly regress point clouds. However, none of them are trained end-to-end to generate the segmentation masks and the point clouds simultaneously. 

Feature tracking~\cite{heinke2019towards,puyol2018fully,qin2018joint} is often used to estimate cardiac motion from cine images which records a cycle of cardiac contraction and relaxation. 
Though tracking methods keep the correspondence between the sequential slices, the shape integrity can't be guaranteed in terms of segmentation. Statistical shape model has the potential to address both the efficiency and shape integrity issues simultaneously since the orders of the points are fixed, ensuring shape integrity and providing boundary correspondences between different frames. 

To address all the issues aforementioned, in this paper, we propose an \textit{end-to-end deep statistical shape model} for image segmentation with both \textit{shape integrity} and \textit{boundary correspondence} preserving: 1) neural network is adopted to extract complex features given the images and learn the shape parameters (both affine and deformation), and
2) the learned parameters are used to warp the mean shape (point cloud) into target shapes which align with the original images. As the final segmentation masks are obtained from the regressed point clouds, small offsets of regressed locations might cause significant mask discrepancies. To overcome this issue, we further introduce a \textit{differentiable rendering layer} to incorporate the mask supervision into the framework to learn better shape parameters. The contributions of this paper can be summarized as follows:
\begin{itemize}
    \item We present a new framework of segmenting myocardium preserving both shape integrity and boundary correspondence in the context of a deep statistic shape model. The proposed framework can predict both the affine and deformation parameters simultaneously. The parameters as well as the mean shape point cloud can then be used to generate target point clouds.
   \item Besides regressing the point clouds, we introduce a differentiable rendering layer to render the predicted point clouds to segmentation masks to incorporate mask supervision, resulting in more accurate transformation parameters, point clouds and masks.
   \item The proposed method is an end-to-end framework, which generates the point clouds and segmentation masks simultaneously, and it is also applicable to other segmentation tasks with specific shape constraints.
\end{itemize}

\section{Related Works}
\textbf{Statistical Shape Model.}
Numerous shape-based segmentation methods have been proposed during the last few decades. For example, the point distribution model (PDM) proposed by Cootes et al.~\cite{cootes1995active} could extract shape variations from a set of training images, whose shapes are represented by point sets. The covariance matrix of these point sets could be calculated from the proposed PDM, and then the most significant eigenvectors of the covariance matrix are used to denote the shape variations. Staib et al.~\cite{staib1992boundary} propose a parametric model which is based on the elliptic Fourier decomposition of the boundary, and then the distributions of the Fourier coefficients are used to learn the shape knowledge. 
Jain et al.~\cite{jain1996object} also propose a general object localization and retrieval scheme based on object shape using deformable templates.

In the deep learning era, researchers have tried to combine traditional shape models with deep convolutional neural networks. Tilborghs et al.~\cite{tilborghs2020shape} propose to learn the shape coefficients of the statistical shape models directly given the CMR image, while Attar et al.~\cite{attar20193d} try to use both image and patient meta-data to learn the shape parameters. Milletari et al.~\cite{milletari2017integrating} propose to use the learned shape parameters to regress the shape locations, and the loss is imposed on the final locations of the points. 
Instead of learning the shape parameters or regressing point locations, we propose an end-to-end framework to learn the affine and deformation parameters simultaneously, and the final loss function is imposed on both the regressed point clouds and the rendered masks.

\noindent \textbf{Shape-Constrained Deep Image Segmentation.}
Incorporating shape constraints into Deep Neural Networks (DNNs) for image segmentation to ensure anatomically consistent results have been extensively explored. 
Cheng et al.~\cite{cheng2016active} adopt an Active Appearance Model (AAM) to refine segmentations from coarse prostate segmentations. Painchaud et al.~\cite{painchaud2020cardiac} propose an anatomical correct segmentation method by identifying the anatomically implausible results and warping them toward the closest anatomically valid shape. TeTrIS~\cite{lee2019tetris} explicitly enforces shape constraints by restricting the model to obtain segmentation through deformations of a given shape prior. We also establish shape priors which is similar to~\cite{lee2019tetris}, while the established template of the proposed method is based on statistical shape model, with both shape integrity and boundary correspondence preserving.

An alternative category of approaches to integrating shape priors into network-based segmentation are topology based methods~\cite{hu2019topology,clough2019explicit}. The key idea is that, by comparing the topological features of the predicted segmented results and their corresponding ground truth, a differentiable loss function~\cite{edelsbrunner2000topological} could be derived to be incorporated into the training of the deep neural networks. 

\noindent \textbf{Mesh Rendering.}
While generating polygons from point clouds (2D or 3D) is straightforward, the process is non-differentiable. Neural mesh renderer~\cite{kato2018neural} has been proposed to map meshes to images, which is differentiable. It is achieved by sampling in a smooth manner. Also, Loper et al.~\cite{loper2014opendr} propose to achieve this goal by approximating the gradient based on image derivatives. All these differentiable renderers are able to backpropagate the gradients from the rendered images/masks back to the vertices/faces of original meshes.

Other differentiable rendering methods have also been proposed, including perhaps the earliest mesh renderer~\cite{smelyansky2002dramatic}, non-mesh differential renders~\cite{insafutdinov2018unsupervised} and the view-based renderer of~\cite{eslami2018neural}. In this paper we focus on 2D image segmentation since 3D dynamic MRI is not yet widely available and we employ the mesh renderer of~\cite{kato2018neural}.

\section{Method}
First, we define the image and the canonical template domain. In the image domain, the imaged heart cross-section varies in location, orientation, scale and deformation. In the canonical template domain, the location/orientation/scale variances are removed by affine image registration, with only the deformation variances left.
Our initial goal is to generate the ground truth point clouds which align with the original images (i.e., in the image domain).   

We propose to use statistical shape model for image segmentation with both shape integrity and boundary correspondence preserving, which can be represented as:
\begin{equation}
\label{statistical}
    P = \theta (P_m + \beta * C_\beta)
\end{equation}
where $P_m$, $P$ denote the pre-defined mean and target point cloud (point cloud aligns with the original image), respectively. $\beta$ is the shape deformation parameter. $C_\beta$ is the principal component matrix, which is composed of the top $\beta^d$ (dimension of $\beta$) eigenvectors computed from the shape space. Additionally, $\theta$ is the affine transformation parameter which transforms a shape from the canonical template domain to the image domain. 

In our case, we use a fixed number of points to denote a shape. Adjusting $\beta$, we could obtain deformed shapes by adding shape variances to the mean shape. The affine transformation is then applied to transform the deformed shape in the canonical template domain to the image domain. 

\begin{figure*}[t]
\centering
\includegraphics[width=1.5\columnwidth]{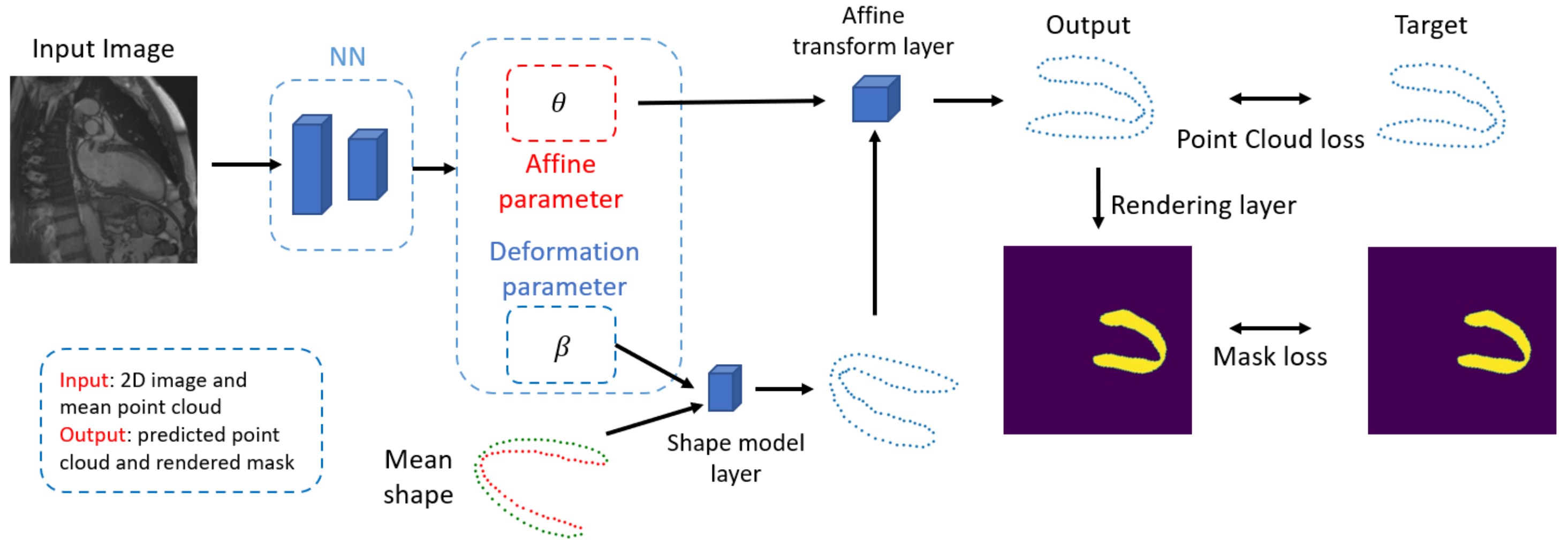}
\caption{Illustration of fitting mean shape point cloud (left down) to both the target point cloud and target mask. Besides the regression loss $L_{i}^{point}$ on the point clouds, we introduce another differentiable rendering layer and incorporate the mask loss $L_{i}^{mask}$.}
\label{method}
\end{figure*}

Fig.~\ref{method} illustrates the overall idea of warping the mean shape point cloud to a point cloud that fits the input image. Given an image, our idea is to find the transformation parameters: $\theta$ and $\beta$. With $\theta$ and $\beta$, the target point cloud which aligns with the input image can be inferred based on Eq.~(\ref{statistical}). 
Traditionally, active shape models or active appearance models are utilized to find these parameters by optimizing loss functions such as least square error. In contrast to these traditional methods, we propose to learn the parameters $\theta$ and $\beta$ in a fully-supervised learning manner. Also, we introduce a differentiable rendering layer to incorporate the mask supervision.

Next we will present our method by four separate parts: data preparation, establishing statistical shape model, training the deep neural networks and introducing the differentiable rendering layer.

\begin{algorithm}[t]
\caption{Point Cloud Generation Workflow}
\label{alg:registration}
\textbf{Input}: total $N$ image and their mask pairs $(I_{i}, M_{i})$\\
\textbf{Output}: image, mask, point clouds which cover the original mask and point clouds in the canonical template domain, quadruples $(I_{i}, M_{i}, P_{i}, P_{i}^{c})$
\begin{algorithmic}[1] 
\STATE Choose the first image and its mask pair $(I_{1}, M_{1})$ as fixing image and mask

\STATE Generate the mean shape ${M_{mean}}$ via \textit{Generalized Procrustes Analysis}
\STATE Generate mean point cloud $P_{mean}$ from obtained mean shape mask ${M_{mean}}$ 
\FOR {$i$ from 1 to $N$}
\STATE Deformation to obtain the point clouds in the canonical template domain $P_i^{c} = P_{mean} \times f(M_{mean} \rightarrow M_{i}^{c})$ 

\COMMENT{$M_{i}^{c}$ denotes the $i$-th mask in canonical template domain}
\STATE Do inverse affine transformation to transform the point clouds in canonical template domain $P_{i}^{c}$ to target point clouds in image domain $P_{i} = P_{i}^{c} \times f(M_{i}^{c} \rightarrow M_{i})$ 

\COMMENT{$M_{i}$ denotes the $i$-th shape in image domain}
\ENDFOR

\COMMENT{The for loop is the inverse transformation to obtain point clouds in image domain}
\STATE \textbf{return} a set of ($N$) quadruples $(I_{i}, M_{i}, P_{i}, P_{i}^{c})$ 
\end{algorithmic}
\end{algorithm}

\subsection{Data Preparation}
Given the 2D original images and their corresponding masks, we generate the ground truth point clouds first. Then the statistical shape model can be established based on the generated point clouds.  

\myparagraph{Point Cloud Generation Workflow.} The details of the point cloud generation process are presented in Alg.~\ref{alg:registration}, where $f$ is the registration operation, and $\times$ indicates the transformation operation. More details are included in the supplementary material.

The ground truth point clouds are obtained by performing forward (image registration) and backward (generate point clouds) operations separately. For the forward operation, we align all the masks in the canonical template domain by removing the location/orientation/scale variances and then compute the mean shape, resulting in mean point cloud. Affine image registration includes two parts: registration initialization and refined affine image registration. We implement the initialization by aligning landmark points. For the backward operation, the inverse transformation (the for loop in Alg.~\ref{alg:registration}) is used to reverse the mean point cloud ${P_{mean}}$ back to image domain (target point clouds $P_{i}$). 


\subsection{Establishing the Statistical Shape Model}
\label{sec:statistical_shape_model}
Point clouds generated via Alg.~\ref{alg:registration} are used in establishing statistical shape model and the subsequent training step. Note that inaccurate point cloud samples are excluded.
Besides the point clouds in the image domain ($P_i$), we also have the corresponding point clouds in the canonical template domain $(P_{i}^{c})$, with location/scale/orientation variances removed and only deformation variations left. We compute the mean point cloud over the training set $P_{mean}^{train}$ in the canonical template domain:
\begin{equation}
    P_{mean}^{train} = \frac{1}{K}\sum_{i=1}^{K} P_i^{c}
\end{equation}
where $K$ is the number of training samples. $P_m$ in Eq.~(\ref{statistical}) is set as $P_{mean}^{train}$. Intuitively, the sequence of $P_i^{c}$ captures all the shape variances in the canonical template domain of the training set. Then, PCA is applied on all these shapes to extract the principal modes of variations by computing the eigenvectors $C=[\phi_1, \phi_2,...,\phi_l]$ and eigenvalues $\Lambda =diag(\lambda_1,\lambda_2,...,\lambda_l) $ of the shape covariance matrix. $l$ is dimension of the shape representation.


\subsection{Training the Deep Neural Network}

We have a set of data triplets: image, mask and point cloud in image domain $(I_{i}, M_{i}, P_{i})$. We can find affine parameter $\theta$ as well as deformation parameter $\beta$ to warp the mean shape to the image domain given an image. 

Fig.~\ref{method} is an illustration of the proposed pipeline. The mean shape point cloud and the matrix $C$ are obtained from statistical shape model part, which are used as prior knowledge.

Given an image, a neural network is adopted to predict its affine parameter $\theta$ and deformation parameter $\beta$. Recall Eq.~(\ref{statistical}), these predicted parameters are then used to warp the mean point cloud to the one in the image domain (target point cloud). The deformation parameter $\beta$ is first used to warp the mean point cloud to canonical template domain,
\begin{equation}
\label{deformation}
    \hat{P^{c}} = P_m + \beta * C_\beta
\end{equation}
where each dimension of $\beta$ denotes the weight of the corresponding principal component. The dimension of $\beta$ ($\beta^d$) is a hyperparameter, depending on how many principal components are used to deform the shapes. The choosing of $\beta^d$ will be discussed in the ablation study section. $C_\beta$ has a dimension of $\beta^d \times l$, and both $\hat{P^{c}}$ and $P_{m}$ have a dimension of $l$ (dimension of a shape representation).

Once we have the deformed shapes, the affine parameter $\theta$ is then used to transform the deformed shapes/point clouds to the original image domain (target shapes/point clouds):
\begin{equation}
\label{affine}
    \hat{P} = \theta \hat{P^{c}}
\end{equation}
where $\theta$ denotes the affine parameter, with a dimension of $2 \times 3$. More formally, we first reshape $\hat{P^{c}}$ obtained from Eq.~(\ref{deformation}) to $2 \times T$ $(l=2T)$, and then the pointwise affine transformation is
\begin{equation}
\left( \begin{array}{c} \hat{x_{j}} \\ \hat{y_{j}} \end{array} \right) = 
\left[ \begin{array}{ccc}
\theta_{11} & \theta_{12} & \theta_{13} \\ \theta_{21} & \theta_{22} & \theta_{23} \end{array} \right]
\left( \begin{array}{c} \hat{x_{j}^{c}} \\ \hat{y_{j}^{c}} \\ 1 \end{array} \right)
\label{affine_tran}
\end{equation}
where $\hat{P}$ is represented as $(\hat{x_1}, \hat{y_1}; \hat{x_2}, \hat{y_2}; ...; \hat{x_T}, \hat{y_T})^{T}$, and $\hat{P^{c}}$ is represented as $(\hat{x_1^{c}}, \hat{y_1^{c}}; \hat{x_2^{c}}, \hat{y_2^{c}}; ...; \hat{x_T^{c}}, \hat{y_T^{c}})^{T}$, both have a dimension of $2 \times T$. Here, $T=l/2$ is the number of fixed points to represent the shape, and $j=1,2, ..., T$ in Eq.~(\ref{affine_tran}).


We formulate the problem as a regression task, and the ground truth point clouds are used to supervise the training. The regression loss in terms of the point clouds for sample $i$ could be formulated as RMSE loss:
\begin{equation}
    L_{i}^{point} = \sqrt{\frac{1}{T}\sum_{t=1}^{T} [(\hat{x_{t}} - x_{t})^{2} + (\hat{y_{t}} - y_{t})^{2}]}
    \label{regress}
\end{equation}
where $\hat{x_{t}}, \hat{y_{t}}, x_{t}, y_{t}$ denote the predicted coordinates of $x$ axis, $y$ axis, and ground truth coordinates of  $x$ axis, $y$ axis for sample $i$, respectively. For simplicity, we ignore the subscript $i$, denoting the sample index, in this section.

\subsection{Differentiable Rendering Layer}
The search space of the parameters ($\beta$ and $\theta$) is large, making it difficult to learn reasonable $\beta$ and $\theta$. Additionally, the RMSE loss only constrains the distance between the predicted and ground truth point clouds. Even with small point cloud differences (i.e., small RMSE loss), the masks generated by the predicted point clouds could still misalign with the original masks (i.e., low Dice score). This motivates us to incorporate the mask supervision into the framework to learn better shape parameters, thus more accurate point clouds and masks. 

Intuitively we consider to convert the predicted point clouds to polygons/segmentation masks.
The key issue here is the \textit{differentiability} so that the neural networks can do back propagation and learn. Our solution is, by introducing the \textit{differentiable rendering layer}, the predicted point clouds are rendered to binary masks and then we can incorporate the segmentation loss. The proposed rendering layer perfectly addresses the differentiability issue.

\begin{figure}[t]
\centering
\includegraphics[width=0.7\columnwidth]{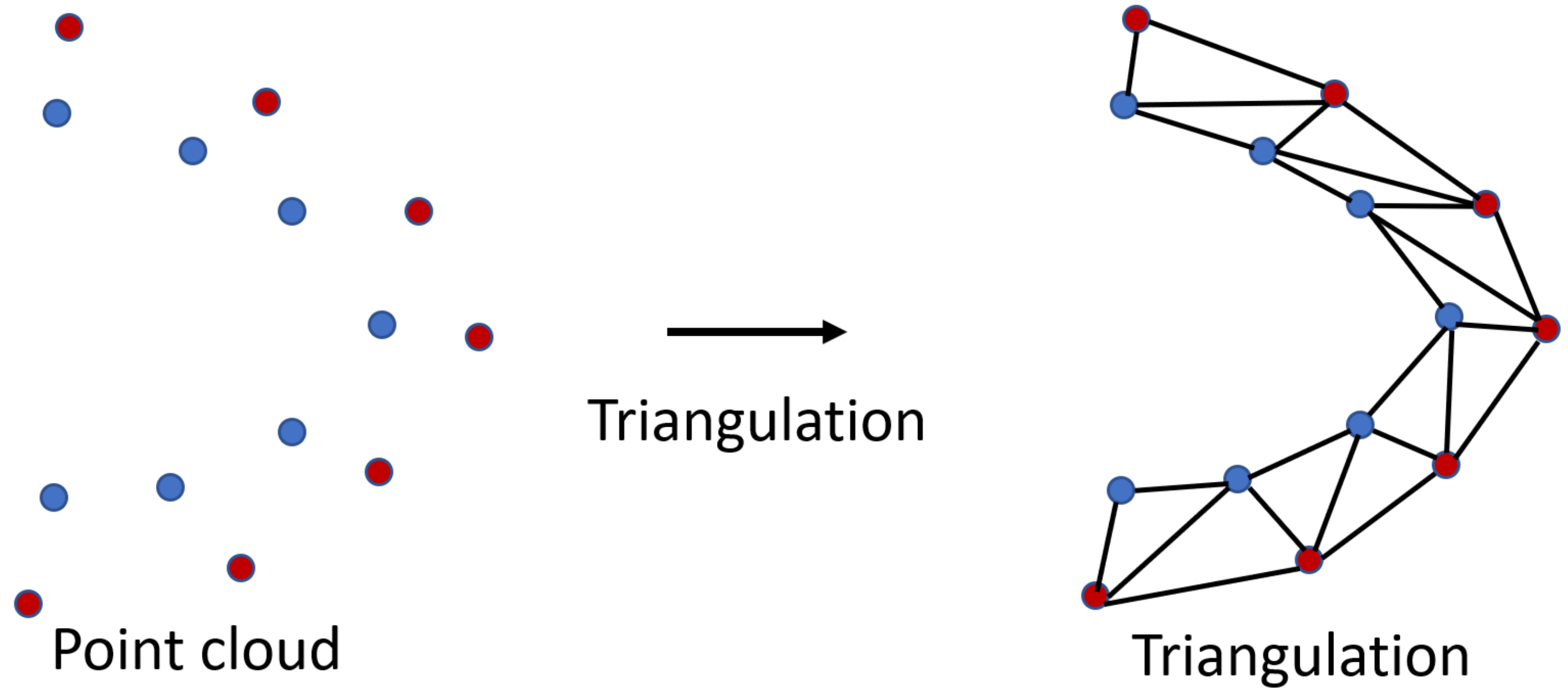}
\caption{Illustration of triangulation. Left is the point cloud, with different colors denoting the points covering the inside (blue) /outside (red) boundary. The bottom point of the inside boundary is indexed as 0.}
\label{triangulation}
\end{figure}

\myparagraph{Triangulation.} The shape of LV myocardium on the long-axis view is concave (Fig.~\ref{teaser}), so we can't simply use Delaunay triangulation as in~\cite{gur2019end}, which results in convex polygons. In our case, we use $T$ points to denote the shape of myocardium, with $T/2$ points ($0, 1, 2, ..., T/2-1$) covering the inside boundary and the rest $T/2$ points ($T/2, T/2+1, T/2+2, ..., T-1$) covering the outside boundary. Instead of using Delaunay triangulation, we formulate the triangles/faces $F$ ourselves. Specifically, the indexes of the triangulated faces are constructed as follows: 
\begin{equation}
\begin{aligned}
  &  \Scale[0.8]{\{0, 1, (T/2)\}, \{1, 2, T/2+1\}, ... , \{T/2-2, T/2-1, T-2\}}\\
    &  \Scale[0.8]{\{T/2, T/2+1, 1\}, \{T/2+1, T/2+2, 2\}, ... , \{T-2, T-1, T/2-1\}}\\
\end{aligned}
\end{equation}
resulting in $(T-2)$ triangles/faces in total. The triangulation process is illustrated in Fig.~\ref{triangulation}.

\myparagraph{Rendering.} 
Given the vertices and the triangulated faces mentioned above, the neural renderer returns a polygon mask, with all pixels inside the faces $F$ are ones, and zeroes otherwise:
\begin{equation}
    \hat{M_i} = R(\hat{P_i}, F)
\end{equation}
where $\hat{P_{i}}$ and $\hat{M_i}$ are the regressed point clouds and rendered segmentation mask for sample $i$, respectively. This mask is almost binary, except for the boundary pixels where interpolation occurs. As the used renderer~\cite{kato2018neural} originally works in 3D space, we project all points to 3D by setting their $z$ axis to 1. As the used renderer is differentiable, we are able to derive an additional loss based on the mask discrepancy and guide the training of the neural networks.

\subsection{Incorporating into End-to-End Training}
Besides the regression loss in terms of the predicted point clouds, by introducing the differentiable rendering layer, the additional mask loss is defined as follows: 
\begin{equation}
    L_{i}^{mask} = dice(\hat{M_{i}}, M_i)
\end{equation}
where $\hat{M_{i}}$, $M_{i}$ denote the rendered and the ground truth masks for sample $i$, respectively. 
And the final loss function is defined as:
\begin{equation}
\label{total}
    L_{i}^{total} = L_{i}^{point} + \delta L_{i}^{mask}
\end{equation}
where $\delta$ is a balanced term.

\section{Experiments}
\myparagraph{Datasets.} 
We validate the effectiveness of the proposed method on three anatomy views from two datasets: 2-chamber and 4-chamber views are from the public cine MRI dataset~\footnote{https://www.kaggle.com/c/second-annual-data-science-bowl/data}. The 3-chamber view is from a private dataset collected by ourselves. All the masks are manually annotated by ourselves. For the public dataset, each cine is from one patient with 30 frames. After the preprocessing and data cleaning procedure, we have 4848, 4031, 6112 samples for 2-chamber, 3-chamber and 4-chamber views, respectively. For all the experiments, we use $80\%$ of the total set as the training set, and the rest $20\%$ as validation set. All the results (mean and standard deviation) are reported on the validation set. There are no patient overlaps between the training and the validation. 


\myparagraph{Baselines.} We choose two popular and general segmentation models, fully convolutional network (FCN)~\cite{long2015fully} and U-Net architecture~\cite{ronneberger2015u}, as baselines. Another contour-based methods~\cite{chen2019learning} is also used as baseline in this paper. For completeness, we compare the proposed method with existing shape based approaches~\cite{lee2019tetris} (TeTrIS), which also enforces shape consistency. To further demonstrate the efficacy of the proposed method, we include one of the SOTA cardiac motion estimation/tracking methods~\cite{qin2018joint} as baseline. 

\myparagraph{Evaluation Metrics.} The Dice similarity coefficient (Dice)~\cite{zou2004statistical} measures the overlapping regions between predicted and ground truth masks. 
The Hausdorff distance (HD)~\cite{huttenlocher1993comparing} calculates the maximum distance between these two boundaries. We also use number of connected components (CC)~\cite{dillencourt1992general} of the segmentation results as a metric.
Besides these traditional quantitative image segmentation evaluation metrics, we also use checkbox to indicate if the segmentation method is shape integrity or boundary correspondence preserving. 

\begin{table*}[ht]
\centering
\begin{center}
\small
\caption{Quantitative results for different models on several benchmark datasets.}
\label{table:quan}
\begin{tabular}{ccccccc}
\hline
Dataset & Method & Integrity & Corre. & Dice $\uparrow$ &  HD (mm) $\downarrow$ & CC\\
\hline\hline
\multirow{6}{*}{\textbf{2-chamber}} &
\textbf{FCN}~\cite{long2015fully} &$\times$ & $\times$ & \textbf{0.827 $\pm$ 0.102} & 3.106 $\pm$ 0.694 & 1.105 $\pm$ 0.698 \\

~ & \textbf{U-Net}~\cite{ronneberger2015u} & $\times$ & $\times$  & \textbf{0.827 $\pm$ 0.104} & 3.052 $\pm$ 0.701 & 1.345 $\pm$ 0.638 \\
~ & \textbf{Contour}~\cite{chen2019learning}& $\times$ & $\times$  & 0.816 $\pm$ 0.123 & 2.950 $\pm$ 0.668 & 1.616 $\pm$ 0.672 \\
~ & \textbf{TeTrIS}~\cite{lee2019tetris}& $\times$ & $\times$  & 0.816 $\pm$ 0.103 & 3.834 $\pm$ 0.621 & 1.917 $\pm$ 1.048\\
~ & \textbf{Tracking}~\cite{qin2018joint} & $\times$& $\surd$  & 0.820 $\pm$ 0.084 & 2.873 $\pm$ 0.530 & 1.210 $\pm$ 0.597\\
 ~ & \textbf{Ours} &$\surd$ & $\surd$ & 0.804 $\pm$ 0.093 & \textbf{2.152 $\pm$ 0.431} & \textbf{1} \\

\hline
\multirow{6}{*}{\textbf{3-chamber}} &
\textbf{FCN}~\cite{long2015fully} & $\times$ & $\times$  & 0.857 $\pm$ 0.050 & 2.822 $\pm$ 0.468 & 1.365 $\pm$ 0.976\\

~ & \textbf{U-Net}~\cite{ronneberger2015u} & $\times$ & $\times$  & 0.864 $\pm$ 0.065 & 2.732 $\pm$ 0.540  & 1.486 $\pm$ 0.848\\

~ & \textbf{Contour}~\cite{chen2019learning}& $\times$ & $\times$  & 0.858 $\pm$ 0.048 & 2.477 $\pm$ 0.500 & 1.299 $\pm$ 0.610 \\
~ & \textbf{TeTrIS}~\cite{lee2019tetris}& $\times$ & $\times$  & \textbf{0.871 $\pm$ 0.053} & 3.758 $\pm$ 0.608 & 1.547 $\pm$ 1.345\\
~ & \textbf{Tracking}~\cite{qin2018joint} & $\times$ & $\surd$  & 0.840 $\pm$ 0.117 & 3.143 $\pm$ 0.643 & 1.349 $\pm$ 0.850 \\
~ & \textbf{Ours} &$\surd$& $\surd$ & 0.840 $\pm$ 0.052 & \textbf{1.938 $\pm$ 0.226} & \textbf{1} \\

\hline
\multirow{6}{*}{\textbf{4-chamber}} &
\textbf{FCN}~\cite{long2015fully} & $\times$ & $\times$  & 0.840 $\pm$ 0.080 & 2.623 $\pm$ 0.493 & 1.330 $\pm$ 0.772\\

~ & \textbf{U-Net}~\cite{ronneberger2015u} & $\times$ & $\times$  & \textbf{0.873 $\pm$ 0.088} & 2.589 $\pm$ 0.795 & 1.273 $\pm$ 0.632\\

~ & \textbf{Contour}~\cite{chen2019learning}& $\times$ & $\times$  & 0.864 $\pm$ 0.076  & 2.345 $\pm$ 0.312 & 1.363 $\pm$ 0.651 \\
~ &\textbf{TeTrIS}~\cite{lee2019tetris}& $\times$ & $\times$  & 0.870 $\pm$ 0.049 & 3.565 $\pm$ 0.462 & 1.498 $\pm$ 1.505 \\
~ &\textbf{Tracking}~\cite{qin2018joint} & $\times$ & $\surd$  & 0.816 $\pm$ 0.136 & 3.137 $\pm$ 0.854 & 1.481 $\pm$ 1.057\\
~ &\textbf{Ours} & $\surd$ & $\surd$& 0.833 $\pm$ 0.051 & \textbf{1.847 $\pm$ 0.268} &\textbf{1}\\
\hline
\vspace{-.3in}
\end{tabular}
\end{center}
\end{table*}

\begin{figure*}[ht]
\centering

\subfigure{
\includegraphics[width=0.12\textwidth, angle =180]{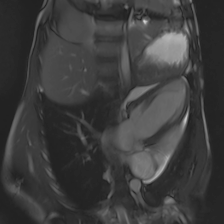}}
\hspace{-.08in}
\subfigure{
\includegraphics[width=0.12\textwidth, angle =180]{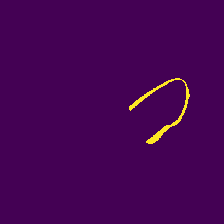}}
\hspace{-.08in}
\subfigure{
\includegraphics[width=0.12\textwidth, angle =180]{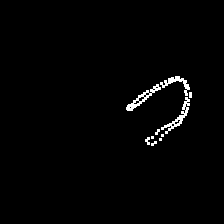}}
\hspace{-.08in}
\subfigure{
\includegraphics[width=0.12\textwidth, angle =180]{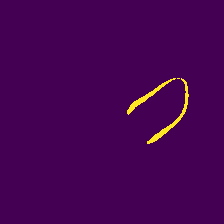}}
\hspace{-.08in}
\subfigure{
\includegraphics[width=0.12\textwidth, angle =180]{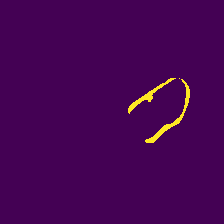}}
\hspace{-.08in}
\subfigure{
\includegraphics[width=0.12\textwidth, angle =180]{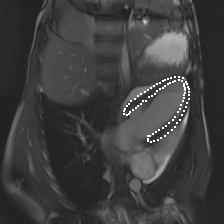}}
\hspace{-.08in}
\subfigure{
\includegraphics[width=0.12\textwidth, angle =180]{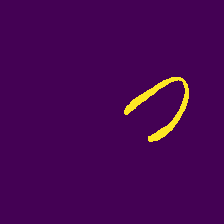}}

\vspace{-.14in}
\subfigure{
\includegraphics[width=0.12\textwidth, angle =90]{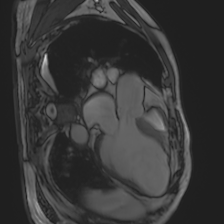}}
\hspace{-.08in}
\subfigure{
\includegraphics[width=0.12\textwidth, angle =90]{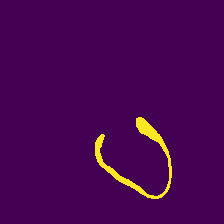}}
\hspace{-.08in}
\subfigure{
\includegraphics[width=0.12\textwidth, angle =90]{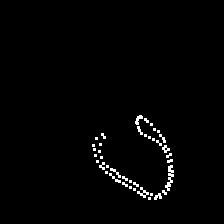}}
\hspace{-.08in}
\subfigure{
\includegraphics[width=0.12\textwidth, angle =90]{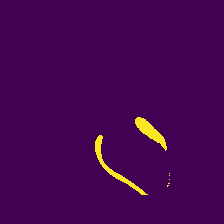}}
\hspace{-.08in}
\subfigure{
\includegraphics[width=0.12\textwidth, angle =90]{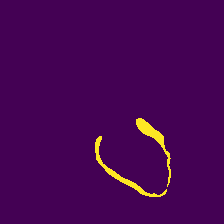}}
\hspace{-.08in}
\subfigure{
\includegraphics[width=0.12\textwidth, angle =90]{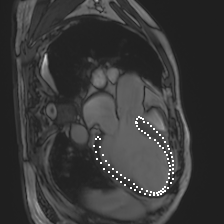}}
\hspace{-.08in}
\subfigure{
\includegraphics[width=0.12\textwidth, angle =90]{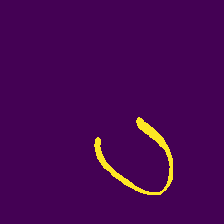}}

\vspace{-.14in}
\subfigure{
\stackunder{\scalebox{1}[-1]{\includegraphics[width=0.12\textwidth]{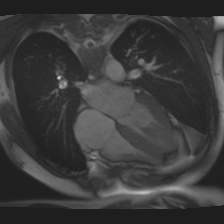}}}{(a) Ori}}
\hspace{-.08in}
\subfigure{
\stackunder{\scalebox{1}[-1]{\includegraphics[width=0.12\textwidth]{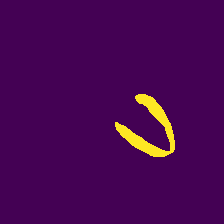}}}{(b) GT}}
\hspace{-.08in}
\subfigure{
\stackunder{\scalebox{1}[-1]{\includegraphics[width=0.12\textwidth]{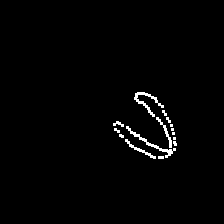}}}{(c) Point}}
\hspace{-.08in}
\subfigure{
\stackunder{\scalebox{1}[-1]{\includegraphics[width=0.12\textwidth]{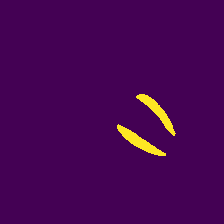}}}{(d) FCN}}
\hspace{-.08in}
\subfigure{
\stackunder{\scalebox{1}[-1]{\includegraphics[width=0.12\textwidth]{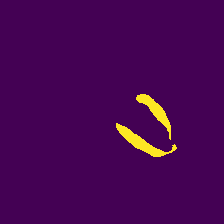}}}{(e) U-Net}}
\hspace{-.08in}
\subfigure{
\stackunder{\scalebox{1}[-1]{\includegraphics[width=0.12\textwidth]{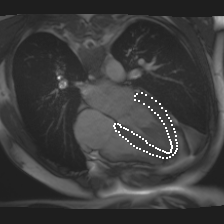}}}{(f) Ours}}
\hspace{-.08in}
\subfigure{
\stackunder{\scalebox{1}[-1]{\includegraphics[width=0.12\textwidth]{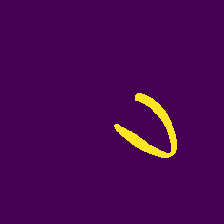}}}{(g) Ours}}

\caption{Qualitative results compared with baselines. Top (2-chamber), middle (3-chamber) and bottom (4-chamber) rows show three representative cases. From left to right: (\textbf{a}) original image, (\textbf{b}) gt mask, (\textbf{c}) gt point clouds, (\textbf{d}) mask generated by FCN, (\textbf{e}) U-Net, (\textbf{f}) point cloud generated by the proposed method overlaid with the original image, and (\textbf{g}) mask generated by the proposed method.}
\vspace{-.05in}
\label{quan}
\end{figure*}

\begin{figure*}[ht]
\centering

\subfigure{
\scalebox{1}[-1]{\includegraphics[width=0.14\textwidth]{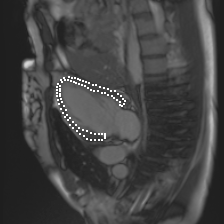}}}
\hspace{-.08in}
\subfigure{
\scalebox{1}[-1]{\includegraphics[width=0.14\textwidth]{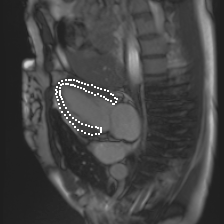}}}
\hspace{-.08in}
\subfigure{
\scalebox{1}[-1]{\includegraphics[width=0.14\textwidth]{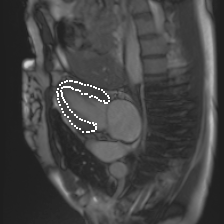}}}
\hspace{-.08in}
\subfigure{
\scalebox{1}[-1]{\includegraphics[width=0.14\textwidth]{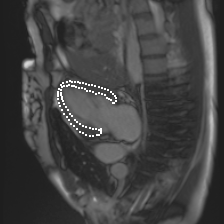}}}
\hspace{-.08in}
\subfigure{
\scalebox{1}[-1]{\includegraphics[width=0.14\textwidth]{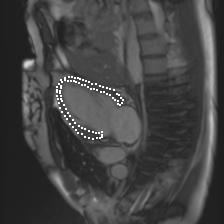}}}
\hspace{-.08in}
\subfigure{
\scalebox{1}[-1]{\includegraphics[width=0.14\textwidth]{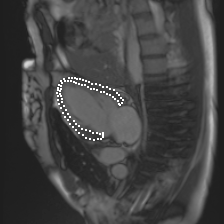}}}

\vspace{-.12in}
\subfigure{
\scalebox{1}[-1]{\includegraphics[width=0.14\textwidth]{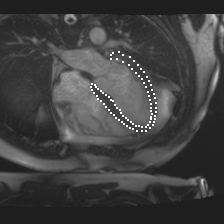}}}
\hspace{-.08in}
\subfigure{
\scalebox{1}[-1]{\includegraphics[width=0.14\textwidth]{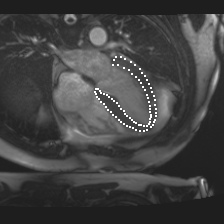}}}
\hspace{-.08in}
\subfigure{
\scalebox{1}[-1]{\includegraphics[width=0.14\textwidth]{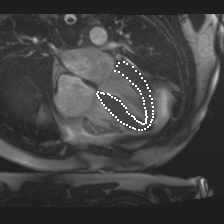}}}
\hspace{-.08in}
\subfigure{
\scalebox{1}[-1]{\includegraphics[width=0.14\textwidth]{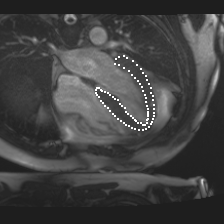}}}
\hspace{-.08in}
\subfigure{
\scalebox{1}[-1]{\includegraphics[width=0.14\textwidth]{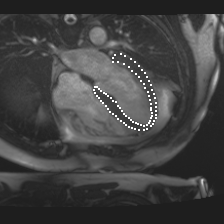}}}
\hspace{-.08in}
\subfigure{
\scalebox{1}[-1]{\includegraphics[width=0.14\textwidth]{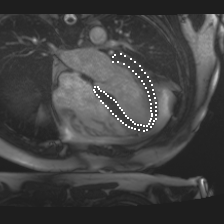}}}



\caption{The predicted point clouds overlay on the original images for a complete cycle of cardiac contraction and relaxation. Example frames from cine MRI show the heart motion starting from relaxation to contraction and then back to relaxation. Top (2-chamber) and bottom (4-chamber) rows show two representative cases. The video version of such cycles are included in the supplementary material.}
\vspace{-.1in}
\label{cycle}
\end{figure*}

\myparagraph{Implementation Details.}
We use official Resnet34 as backbone to extract features, and three fully connected layers are attached to the end to learn the affine and deformation parameters. The network uses Adam for optimizing the loss function (Eq.~(\ref{total})) with the learning rate of 0.001 and batch size of 256. To mitigate the overfitting issue, we adopt popular data augmentation techniques, such as flip (Up-Down, Left-Right), rotation, and translation. We apply the same augmentation to the point cloud. Note that the point cloud rotation centers are the same as the corresponding image rotation centers. In all of our experiments, we set $\beta^d = 30$ and $\delta = 0.5$. Additionally, each dimension of deformation parameter $\beta$ is restricted to $\beta_j \leq 3\sqrt\lambda_j$ to capture the 99.7\% of shape variability. For 4-chamber view (6112 images), it takes $\approx$10 hours to train the network. And it takes $\approx$0.1s to inference an image with the size of 224 $\times$ 224. All the experiments are performed on a Tesla V100 GPU (16G), and an Intel(R) Xeon(R) E5-2630 v4 CPU@2.20 GHz.

More details of \textbf{Landmarks Generation} and \textbf{Training Strategy} are included in supplementary material.

\subsection{Results}
\myparagraph{Quantitative Results.}
The quantitative results compared with other baselines are shown in Tab.~\ref{table:quan} and we report both the mean and standard deviations for all the metrics. 
The results are highlighted with bold when they are significantly better (Student's t-test, with a confidence interval of 95\%).

We want to emphasize that compared with the proposed method, TeTrIS~\cite{lee2019tetris} is the only one which also uses shape prior, while none of these segmentation based baselines (FCN~\cite{long2015fully}, U-Net~\cite{ronneberger2015u}, Contour~\cite{chen2019learning} and TeTrIS~\cite{lee2019tetris}) are shape integrity nor boundary correspondence preserving. For the tracking baseline, Tracking~\cite{qin2018joint} is proposed for cardiac motion estimation and segmentation simultaneously. Note that though tracking methods keep the correspondence between the sequential slices, they are not shape integrity preserving. Moreover, in clinical context, instead of pixel-level accuracy, shape integrity and boundary correspondence are more important for downstream tasks, such as cardiac motion estimation and disease diagnosis.

As shown in Tab.~\ref{table:quan}, the proposed method achieves slightly inferior performances in terms of Dice compared to the state-of-the-art segmentation based baselines (FCN~\cite{long2015fully}, U-Net~\cite{ronneberger2015u}, Contour~\cite{chen2019learning} and TeTrIS~\cite{lee2019tetris}). As these segmentation methods are optimized with pixel-wise losses, such as cross entropy or Dice loss, it's not surprising that they achieve impressive Dice scores, which is also a pixel-wise metric, measuring the overlapping regions between predicted and ground truth masks. 

The proposed method achieves significantly better Hausdorff distance (HD) than all other baselines. Hausdorff distance (HD) actually calculates the maximum distance between the predicted and ground truth boundaries, which is sensitive to false positives that are far away from the ground truth boundaries. As the proposed method exploits shape models and regresses the boundary points directly ($L^{point}$), in the inference stage, the model will always generate segmentation masks with reasonable boundaries which are close to the ground truth boundaries, ensuring an upper-bound for HD. On the other hand, all the baselines are optimized with pixel-wise losses, such as cross-entropy or Dice loss. Consequently, all of them will inevitably produce false positive labels which might be far away from the ground truth mask boundaries. Even a small set of false positive labels which are far away from the ground truth boundaries will cause big Hausdorff distances.

For all the methods, we threshold the obtained mask at 0.5 to generate the binary mask. And then a 8-connected component analysis is performed to obtain the number of connected components. Ideally, all segmentations should have only one connected component, which is one of the main advantages of the proposed method. Though the baseline methods might achieve quite good numbers in terms of the other segmentation metrics, all of them fail to produce single connected component mask without post processing. Generally speaking, compared with the baseline methods, \textit{the proposed method achieves shape integrity and boundary correspondence with sacrificing segmentation accuracy a little bit}.

\myparagraph{Qualitative Results.} The qualitative results comparing with other methods are shown in Fig.~\ref{quan}. Compared with FCN (Fig.~4(d)) and U-Net (Fig.~4(e)), the proposed method could generate masks with shape integrity (Fig.~4(g)).

Additionally, the points in the predicted point clouds are ordered, which means there are correspondences between the sequential predicted point clouds. Fig.~\ref{cycle} shows the results for a complete cycle of cardiac contraction and relaxation. The correspondence property will contribute a lot to the downstream analysis, such as motion estimation of myocardium, which further demonstrates the potential of the proposed approach.

\subsection{Ablation Studies}
Here, we conduct several ablation studies to demonstrate the effectiveness of the proposed method and provide a rough guideline of choosing the hyperparameters. Note that all the ablation studies are conducted on 4-chamber dataset. 

\vspace{-.05in}
\setlength{\tabcolsep}{5pt}
\begin{table}[ht]
\begin{center}
\small
\caption{Ablation study for backbones.}
\label{table:backbone}
\begin{tabular}{ccc}
\hline
 Backbone & Dice $\uparrow$ &  HD (mm) $\downarrow$\\
\hline\hline
\textbf{Resnet18} & 0.705 $\pm$ 0.069 & 2.110 $\pm$ 0.256 \\

\textbf{Resnet50} & 0.827 $\pm$ 0.079  & 1.923 $\pm$ 0.296 \\
\textbf{Ours (Resnet34)} & \textbf{0.833 $\pm$ 0.051} & \textbf{1.847 $\pm$ 0.268} \\
\hline
\vspace{-.4in}
\end{tabular}
\end{center}
\end{table}
\myparagraph{Ablation Study for Backbones.} In all our experiments, we use the official Resnet34 as the backbone. Here, we compare the performances between different backbones. Specifically, we also tried Resnet18 and Resnet50. Tab.~\ref{table:backbone} shows the results for different backbones. From the table, we can see that Resnet34 achieves slightly better performances than Resnet50 in both metrics, and both of them perform much better than Resnet18. While compared with Resnet34, Resnet50 is more complex and has more trained parameters (26M vs 22M), which makes the model more difficult to train and converge. Consequently, to balance the performances and model complexity, we choose Resnet34 as our final backbone. 
 
\setlength{\tabcolsep}{4pt}
\begin{table}[ht]
\begin{center}
\small
\caption{Ablation study for loss terms.}
\label{table:loss_ablation}
\begin{tabular}{ccc}
\hline
 Backbone & Dice $\uparrow$ &  HD (mm) $\downarrow$\\
\hline\hline
\textbf{w/o Rendering} ($L_{i}^{point}$) & 0.682 $\pm$ 0.126 & \textbf{1.843 $\pm$ 0.251}  \\
\textbf{Mask loss only} ($L_{i}^{mask}$) & 0.755 $\pm$ 0.097  & 1.875 $\pm$  0.314 \\
\textbf{Ours} & \textbf{0.833 $\pm$ 0.051} & \textbf{1.847 $\pm$ 0.268} \\
\hline
\vspace{-.2in}
\end{tabular}
\end{center}
\end{table}
\myparagraph{Ablation Study for Different Loss Terms.} In order to preserve shape integrity and boundary correspondence, we adopt the shape model to generate point clouds. While based on our observation, as the search space of the learned parameters is quite large, the segmentation accuracy, Dice score, is far from perfect ($L_{i}^{point}$ row in Tab.~\ref{table:loss_ablation}). The reason is that, the loss $L^{point}_{i}$ is a regression loss, and it minimizes the distances between the predicted and ground truth point clouds. While Dice score is a metric of segmentation, and it is very sensitive to the offsets of point clouds. Consequently, we introduce the differentiable rendering layer and incorporate the mask loss as another supervision. 

On the other hand, the performance drops significantly if only the mask loss is used ($L_{i}^{mask}$ row in Tab.~\ref{table:loss_ablation}). As the rendered mask and the proposed mask loss rely heavily on the regressed point clouds, the point cloud regression loss $L_{i}^{point}$ helps to minimize the offsets of point clouds, ensuring to obtain reasonable segmentation masks. The ablation study results in Tab.~\ref{table:loss_ablation} validates the contributions of both loss terms and the proposed rendering layer.

\setlength{\tabcolsep}{4pt}
\begin{table}[ht]
\begin{center}
\small
\caption{Ablation study for number of principal components used for prediction.}
\label{table:pca}
\begin{tabular}{cccc}
\hline
$\beta^d$ & Dice $\uparrow$ &  HD (mm) $\downarrow$\\
\hline\hline

\textbf{10} & 0.376 $\pm$ 0.093 & 2.129 $\pm$ 0.226\\
\textbf{20} & 0.682 $\pm$ 0.114 & 1.956 $\pm$ 0.219\\
\textbf{25} & 0.702 $\pm$ 0.111 & \textbf{1.781 $\pm$ 0.207}\\
\textbf{Ours (30)} & \textbf{0.833 $\pm$ 0.051} & 1.847 $\pm$ 0.268\\
\hline
\vspace{-.3in}
\end{tabular}
\end{center}
\end{table}

\myparagraph{Ablation Study for $\beta^d$.} The number of principal components ($\beta^d$) used to learn the parameters is a pre-defined parameter, and could be roughly determined offline based on the eigenvalues computed on the training set. In our case, the top 30 biggest eigenvalues consist around 99\% energies of the total. Fewer number of principal components used will cause loss of information (in the deformation part/PCA layer), resulting in bad performances (See Tab.~\ref{table:pca}). 

\setlength{\tabcolsep}{4pt}
\begin{table}[ht]
\begin{center}
\small
\caption{Ablation study for loss weights.}
\label{table:delta}
\begin{tabular}{ccc}
\hline
$\delta$ & Dice $\uparrow$ &  HD (mm) $\downarrow$\\
\hline\hline
\textbf{0} & 0.682 $\pm$ 0.126 & \textbf{1.843 $\pm$ 0.251}  \\
\textbf{0.1} & 0.793 $\pm$ 0.061& 1.946 $\pm$ 0.235 \\
\textbf{0.2} & 0.754 $\pm$ 0.072 & 1.909 $\pm$ 0.231 \\
\textbf{1.0} & 0.642 $\pm$ 0.066 & \textbf{1.851 $\pm$ 0.214} \\
\textbf{Ours (0.5)} & \textbf{0.833 $\pm$ 0.051} & \textbf{1.847 $\pm$ 0.268} \\
\hline
\vspace{-.3in}
\end{tabular}
\end{center}
\end{table}
\myparagraph{Ablation Study for $\delta$.} $\delta$ is the balanced term between the point cloud loss and the mask loss, which is critical to achieve good performance. The rationale is that, if $\delta$ is too small, the incorporated mask loss can't fully exert its power to correct the misaligned point clouds. On the other hand, the point cloud loss can be regarded as an initialization of the mask loss. If $\delta$ is too large, the power of point cloud loss decreases, and therefore couldn't provide a good initialization for the mask loss, making the model difficult to converge. Tab.~\ref{table:delta} illustrates the quantitative results of different weights of the mask loss.

\section{Conclusion}

In this paper, we propose a novel end-to-end deep statistic shape model for myocardium segmentation with both shape integrity and boundary correspondence preserving. The proposed framework learns both the affine and deformation parameters simultaneously. Besides the point cloud regression loss, a differentiable rendering layer is introduced to incorporate mask supervision by rendering the predicted point clouds to segmentation masks, which helps to boost the performances. Several ablation studies have been conducted to demonstrate the efficacy of the proposed method and provide a guideline of choosing hyperparameters. Also, the method proposed in this paper is a general framework, and it is also applicable to other segmentation tasks with specific shapes.

{\small
\bibliographystyle{ieee_fullname}
\bibliography{shape_model_arXiv}
}

\clearpage
\section{Appendix}

\appendix
\section{Image registration details}

Fig.~\ref{registration} illustrates the image registration workflow. Three key points/landmarks (one apex and two basal points) are extracted for each image. See second column of Fig.~\ref{registration}. The key points are colored. Specifically, the landmark detection problem is treated as a regression problem where landmark probability map is generated from a U-Net like neural network. Then the landmark location is determined by detecting the maximum value. For multiple landmarks (the number is 3 in our case), the network outputs multiple channels where each landmark is estimated from each channel.
 
Function \textit{cv2.estimateAffinePartial2D}~\footnote{http://amroamroamro.github.io/mexopencv/matlab/cv.estimateAffinePartial2D.html} is used to align these key points (landmark registration). Function \textit{ants.registration}~\footnote{https://antspy.readthedocs.io/en/latest/registration.html} is then used to do refined affine registration and image deformation.

\section{Landmarks Generation} We first extract the boundaries of the computed mean mask (including both interior and exterior boundaries). Then the landmarks are uniformly sampled on the interior and exterior boundaries separately. There are $T/2= 44$ points for interior and exterior boundaries, respectively. Also, the choice of $T$ is a trade-off between accuracy (covering the mask) and computation complexity. We use libraries in Python to generate polygons from the obtained point clouds and compute the Dice scores between the generated polygons and ground truth masks offline. Dice score is used as a metric of accuracy to evaluate the quality of generated point clouds. Separate shape models and networks are trained for each view.

\section{Training Strategy}
The search space of the learned parameters is large, which makes it difficult for the models to converge. Specifically, we remove the differentiable rendering layer first and train the deep neural network with only the point cloud regression loss $L_{i}^{point}$. Once the model is converged, we add the additional mask loss $L_{i}^{mask}$ to finetune the model obtained from the first step to refine the predicted point clouds and masks.

\setcounter{figure}{5}
\begin{figure*}[t]
\centering
\includegraphics[width=0.7\textwidth]{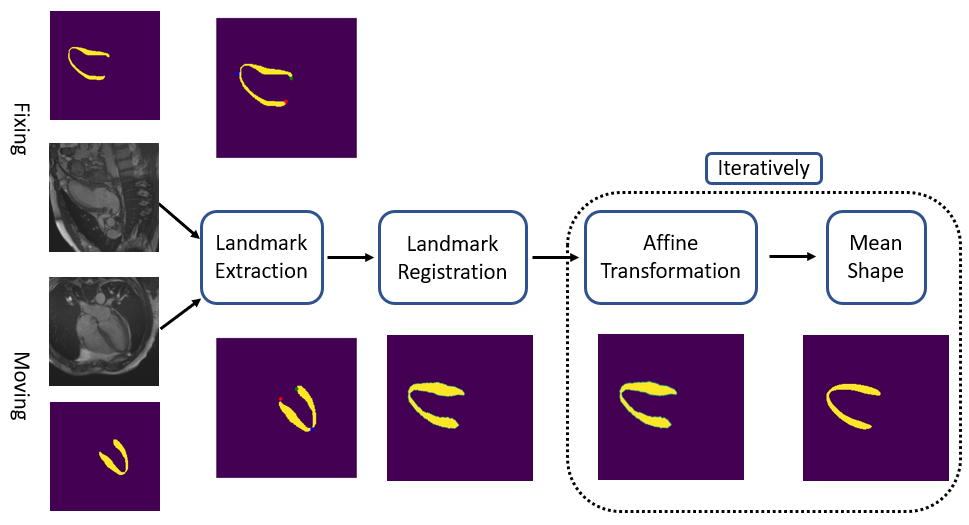}
\caption{Detailed image registration workflow.}
\label{registration}
\end{figure*}

\section{More qualitative results} 
\begin{figure*}[t]
\centering

\subfigure{
\stackunder{\scalebox{-1}[1]{\includegraphics[width=0.13\textwidth]{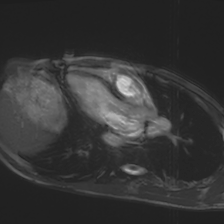}}}{(a)}}
\hspace{-.08in}
\subfigure{
\stackunder{\scalebox{-1}[1]{\includegraphics[width=0.13\textwidth]{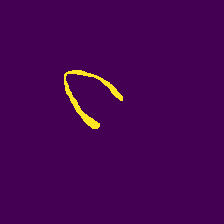}}}{(b)}}
\hspace{-.08in}
\subfigure{
\stackunder{\scalebox{-1}[1]{\includegraphics[width=0.13\textwidth]{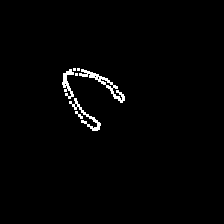}}}{(c)}}
\hspace{-.08in}
\subfigure{
\stackunder{\scalebox{-1}[1]{\includegraphics[width=0.13\textwidth]{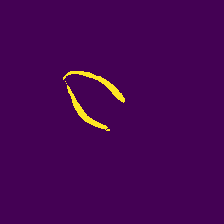}}}{(d)}}
\hspace{-.08in}
\subfigure{
\stackunder{\scalebox{-1}[1]{\includegraphics[width=0.13\textwidth]{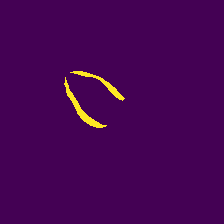}}}{(e)}}
\hspace{-.08in}
\subfigure{
\stackunder{\scalebox{-1}[1]{\includegraphics[width=0.13\textwidth]{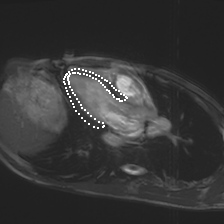}}}{(f)}}
\hspace{-.08in}
\subfigure{
\stackunder{\scalebox{-1}[1]{\includegraphics[width=0.13\textwidth]{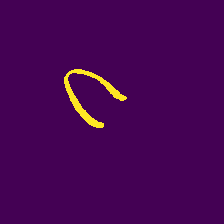}}}{(g)}}

\caption{Qualitative results compared with baselines. From left to right: (\textbf{a}) original image, (\textbf{b}) gt mask, (\textbf{c}) gt point clouds, (\textbf{d}) mask generated by FCN, (\textbf{e}) U-Net, (\textbf{f}) point cloud generated by the proposed method overlaid with the original image, and (\textbf{g}) mask generated by the proposed method.}
\label{supple_quan}
\end{figure*}

More qualitative results comparing with the baseline models are illustrated in Fig.~\ref{supple_quan}.

\section{Illustration of complete cycle of contraction and relaxation} 
\begin{figure*}[ht]
\centering


\subfigure{
\scalebox{1}[-1]{\includegraphics[width=0.15\textwidth]{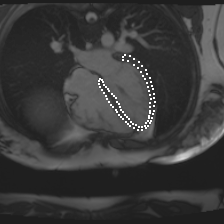}}}
\hspace{-.08in}
\subfigure{
\scalebox{1}[-1]{\includegraphics[width=0.15\textwidth]{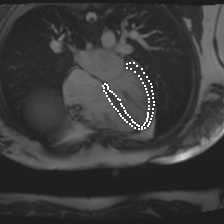}}}
\hspace{-.08in}
\subfigure{
\scalebox{1}[-1]{\includegraphics[width=0.15\textwidth]{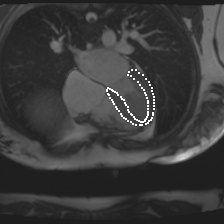}}}
\hspace{-.08in}
\subfigure{
\scalebox{1}[-1]{\includegraphics[width=0.15\textwidth]{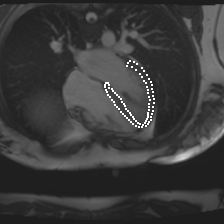}}}
\hspace{-.08in}
\subfigure{
\scalebox{1}[-1]{\includegraphics[width=0.15\textwidth]{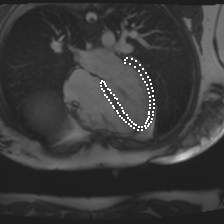}}}
\hspace{-.08in}
\subfigure{
\scalebox{1}[-1]{\includegraphics[width=0.15\textwidth]{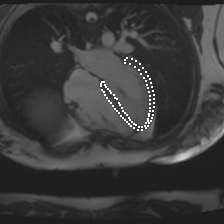}}}

\caption{The predicted point clouds overlay on the original images for a complete cycle of cardiac contraction and relaxation. Example frames from a long-axis cine MRI show the heart motion starting from relaxation to contraction and then back to relaxation.}
\label{supple_cycle}
\end{figure*}

More qualitative results for complete cardiac cycles of contraction and relaxation are illustrated in Fig.~\ref{supple_cycle}. Also, the whole frames for individual patients are made as videos, which are named as patient1.avi, patient2.avi and patient3.avi. These videos are contained in the supplementary material.

\end{document}